\documentclass[10pt,twocolumn,letterpaper]{article}

\usepackage[final]{cvpr}      
\usepackage{amsmath}
\usepackage{amssymb}
\usepackage{booktabs}
\usepackage{url}
\usepackage{mathtools} 

\usepackage{times,graphicx,amsmath,amssymb,booktabs,tabulary,multirow,overpic,xcolor}
\usepackage{colortbl}
\usepackage{makecell}
\usepackage{float}
\usepackage{bbding}

\definecolor{sh_gray}{rgb}{0.84,0.84,0.84}
\definecolor{sh_gray2}{rgb}{1,0.89,0.75}
\definecolor{color3}{rgb}{0.95,0.95,0.95}
\definecolor{color4}{rgb}{0.96,0.96,0.86}
\definecolor{color5}{rgb}{0.90,0.90,0.90}
\usepackage[pagebackref,breaklinks,colorlinks]{hyperref}
\usepackage[capitalize]{cleveref}
\usepackage{multirow}
\crefname{section}{Sec.}{Secs.}
\Crefname{section}{Section}{Sections}
\Crefname{table}{Table}{Tables}
\crefname{table}{Tab.}{Tabs.}
\usepackage{algorithm}
\usepackage{algorithmic}

\begin{document}

\title{Enabling Real-Time Colonoscopic Polyp Segmentation on Commodity CPUs via Ultra-Lightweight Architecture}

\author{%
Weihao Gao$^{1,}$\thanks{Corresponding author. Email: weihaomeva@163.com} , 
Zhuo Deng$^{2}$, 
Zheng Gong$^{2}$, 
Lan Ma$^{2}$\\[0.5em]
$^{1}$School of Computer Science and Artificial Intelligence, Guangdong University of Education\\
$^{2}$Shenzhen International Graduate School, Tsinghua University\\
}
    
\maketitle

\begin{abstract}

Real-time polyp segmentation is essential for early colorectal cancer detection, yet clinical deployment remains blocked by GPU dependency. We introduce the UltraSeg family, a set of CPU-native segmentation models operating below 0.3M parameters. UltraSeg-108K (0.108M) establishes the extreme-compression frontier, while UltraSeg-130K (0.130M) integrates cross-layer lightweight fusion for enhanced multi-center generalization. The architecture replaces parameter-heavy components with grouped multi-rate dilated convolutions and attention-gated cross-layer fusion, achieving real-time throughput on a single CPU core (exceeding 50 FPS at 256×256 and 30 FPS at 352×352) without sacrificing clinical-grade accuracy. Evaluated on seven public datasets, UltraSeg-130K attains Dice scores exceeding 0.8 at both resolutions, substantially outperforming all existing sub-0.3M competitors. Notably, it approaches or exceeds UNet-Medium (7.76M parameters) on zero-shot external validations while using only 1.7\% of its parameters, establishing the first strong baseline for CPU-native real-time polyp segmentation. When scaled to 4.38M parameters, UltraSeg achieves accuracy competitive with heavyweight state-of-the-art models while maintaining an order-of-magnitude parameter advantage, demonstrating that the proposed design principles yield intrinsic representational gains across the entire efficiency spectrum. By delivering the first clinically deployable, CPU-native real-time solution, this work provides an immediately usable tool for resource-limited settings and a reproducible blueprint for real-time medical AI beyond endoscopy. Source code is publicly available.

\end{abstract}

\section{Introduction}
\label{sec:introduction}

Colorectal cancer often arises from the malignant transformation of colorectal polyps, and both its incidence and mortality remain high~\cite{morgan2023global}. Colonoscopy, the primary screening modality for these polyps, allows direct visualization of their morphology and facilitates the resection of malignant lesions~\cite{saraiva2021colorectal,biffi2022novel}. However, the intricate anatomy of the colon frequently causes polyps to be missed. Recent work has explored image-classification or object-detection techniques to assist real-time video interpretation~\cite{ahmad2023evaluation,lalinia2024colorectal,wang2021artificial,li2022performance}, yet polyp segmentation goes further: it supplies pixel-level location and contour information, provides richer visual cues for intra-operative resection, and enables models to learn deeper pixel-wise representations of polyps, thereby boosting the performance of downstream diagnostic tasks~\cite{goh2024risk}. Accurate polyp segmentation is therefore essential for the early diagnosis of colorectal cancer.

Nevertheless, polyp segmentation remains a complex and challenging task. Polyps often exhibit indistinct boundaries and highly variable size and morphology, giving them strong camouflage characteristics~\cite{jong2025gastronet}. Traditional manual segmentation relies on the clinician’s subjective experience and sustained attention, resulting in limited consistency. Early computer-assisted methods predominantly exploited low-level features such as shape, texture, or color; although fast, their accuracy and robustness are insufficient when the contrast between polyps and surrounding mucosa is low~\cite{bernal2012towards}.

In recent years, convolutional neural networks (CNNs)—a cornerstone of deep learning—have drawn substantial attention for polyp segmentation. Among them, U-Net stands out as the most representative architecture; its elegant design and compelling performance have made it a de facto standard in medical image segmentation~\cite{ronneberger2015u}. As deep learning evolves, emerging backbones such as Transformer and Mamba have also been transplanted to this field, yet the accompanying surge in parameters often compromises clinical deployability~\cite{gu2024mamba,dong2021polyp,2021An}.

Accurate polyp segmentation is pivotal for early cancer diagnosis; however, several critical challenges persist in automated workflows. First, inter-scanner variability—stemming from differences in imaging processors and light sources across manufacturers (e.g., Olympus, Pentax, Fujifilm)—renders the same lesion markedly divergent in color, texture, and geometric distortion. Models trained on one device therefore suffer pronounced performance drops when transferred to another. Second, the coexistence of white-light endoscopy (WLE), narrow-band imaging (NBI), blue-laser imaging (BLI), and linked-color imaging (LCI) implies that the same polyp can exhibit drastically different contrast and edge characteristics under distinct spectral modalities, further complicating the extraction of universal features~\cite{jha2024polypdb}.

On the other hand, real-time colonoscopy video processing demands over 30 FPS inference, yet even the lightest published polyp-segmentation networks still carry around 10M parameters and must run on a GPU. This severely restricts deployment in primary hospitals, mobile devices, or embedded endoscopic systems~\cite{tesema2025lgps, PAN2026112101,borgli2020hyperkvasir}. Shrinking a model to a CPU-friendly size while preserving clinically usable accuracy, therefore, remains an open problem.

Current lightweight-colonoscopy research almost universally follows a “top-down” paradigm: start with a ResNet or ViT backbone pretrained on ImageNet, plug it into a U-shaped decoder to obtain a U-Net-like segmenter, and then compress further via pruning, structural redesign, or knowledge distillation~\cite {dong2021polyp,bhattacharya2024polypnextlstm,lin2024polyp}. Yet, even under these performance-preserving constraints, the parameter count can only be pushed around 10M, which is still too heavy for real-time CPU inference.

Unlike penetrative deep-anatomical modalities such as X-ray and MRI, both dermoscopy and colonoscopy operate under visible-light surface imaging. The former targets human skin tissue, whereas the latter targets gastrointestinal mucosal tissue. This shared imaging physics (surface-reflectance optics), coupled with structural task-level correspondence (single-class ROI segmentation, real-time diagnostic demand), establishes the foundation for architectural transfer between the two domains. Driven by the rigid demand for edge-device deployment in dermatology, the dermoscopy community has produced several native ultra-lightweight architectures (e.g., EGE-UNet and LB-UNet, with fewer than 100K parameters)~\cite{xu2024lb,ruan2023ege}. By contrast, no validated sub-100K native design exists in colonoscopy polyp segmentation research. Consequently, dermatology constitutes the sole mature, empirically validated source of ultra-lightweight architectures when establishing a sub-300K baseline for CPU-native real-time inference. This dual alignment at both the imaging and task levels enables dermatology-validated micro-backbones to be transferred via colonoscopy-oriented cross-modal structural redesign and adaptive enhancement, thereby breaking the existing parameter lower bound without resorting to post-hoc compression of cumbersome networks.

Guided by this insight, we establish a radically different "bottom-up" paradigm. We take a proven dermoscopic lightweight model as the seed, eschew traditional compression, and introduce (i) colonoscopy-oriented structural refinements, (ii) an Enhanced Dilated Block for enlarged multi-scale receptive fields, and (iii) a Cross-Layer Lightweight Fusion module for improved cross-center generalization. Every architectural increment is governed by a strict "parametric minimalism" principle: any gain in segmentation performance must be purchased with the smallest possible parameter overhead. In this way, we trade upward only what is essential, achieving a precise equilibrium between capability and deployability.

We present UltraSeg, where "Ultra" denotes the ultralight regime, establishing the first strong baseline for CPU-real-time medical segmentation. Two instantiations are released: UltraSeg-108K (108K parameters), optimized for single-center, single-modality scenarios, delivers over 50 FPS on a single CPU core; and UltraSeg-130K (130K parameters), which augments the backbone with cross-layer attention, is optimized for more complex multi-center, multi-modal clinical deployment scenarios.

These models are evaluated on six public polyp segmentation datasets (CVC-ClinicDB, Kvasir, PolypGen, PolypDB, ETIS-Larib, BKAI-IGH)~\cite{CVC-ClinicDB,jha2020kvasir,ali2106polypgen,jha2024polypdb,son2024polyps,dumitru2023using} and one surgical instrument segmentation dataset (Kvasir-Instrument)~\cite{kvasirinst}. On mixed polyp segmentation datasets, UltraSeg-108K achieves Dice scores of 0.7884 at 256 resolution and 0.8060 at 352 resolution, while UltraSeg-130K attains 0.8038 at 256 resolution and 0.8134 at 352 resolution, substantially outperforming all existing peers under the "extreme-compression" regime.\footnote{We define the "extreme-compression paradigm" as models with $<0.3$M parameters, a threshold empirically determined from our real-time segmentation profiling on commodity single-core CPUs ($>30$FPS).} Notably, external validation on unseen datasets (ETIS-Larib and BKAI-IGH) demonstrates superior robustness, with performance markedly surpassing other lightweight models and approaching or exceeding that of UNet-Medium (7.76M parameters).

Furthermore, scaling UltraSeg-130K to 4.38M parameters (UltraSeg-4.38M) achieves 0.8503 Dice, surpassing UNet-Base with merely 14\% of its parameters and 8\% of its FLOPs, validating the architectural scalability beyond the extreme-compression regime. This work demonstrates that foresighted architectural design, rather than post-hoc compression, can yield high-performance AI ready for extremely resource-constrained environments while maintaining intrinsic representational efficiency across the entire capacity spectrum.

Operating well below the 0.3M-parameter ceiling, the UltraSeg family exceeds 50 FPS on a single core, establishing the current best-performing lightweight solution for real-time colonoscopic segmentation on CPU-only hardware. To ensure clinical utility beyond benchmark metrics, we release not only the model weights but also basic real-time video processing pipelines. This provides a turnkey solution for clinical validation.

To summarize, the main contributions are:

\begin{enumerate}
\item Establishment of the extreme-compression paradigm ($<0.3$M parameters), where UltraSeg-108K and UltraSeg-130K achieve $0.7884$ and $0.8038$ Dice under mixed-dataset training. Rigorous external validation on ETIS-Larib and BKAI-IGH demonstrates superior robustness.
    
\item Demonstration of intrinsic architectural efficiency across the capacity spectrum: scaling from$0.13$M to $4.38$M parameters yields consistent gains, with UltraSeg-4.38M surpassing the vanilla UNet baseline, showcasing the holistic performance gains enabled by structural optimization.

\item Feasibility validation of real-time inference on commodity CPUs and memory-constrained edge devices, offering a turnkey solution for resource-limited clinical settings.
\end{enumerate}

\section{Related Work}
\label{sec:related}

\subsection{Medical image segmentation model}
\label{retable}

The rise of CNNs has radically reshaped medical image segmentation. Among them, UNet—with its encoder-decoder symmetry and skip connections, has become the de facto standard for a wide range of biomedical tasks~\cite{ronneberger2015u}. Numerous successors enrich this backbone with attention mechanisms~\cite{xu2024lb,ruan2023ege}, more elaborate decoders, or multi-scale feature fusion~\cite{niu2024fnexter,dong2021polyp}, yielding steady gains on specific datasets.

In recent years, Transformer architectures have been transplanted into medicine thanks to their capacity for global-context modeling~\cite{dong2021polyp,lin2024polyp}. Swin-Transformer and pyramid vision transformers, for example, excel at capturing multi-scale spatial dependencies~\cite{liu2021swin,wang2021pyramid,wang2022pvt}. Meanwhile, state-space models such as Mamba are gaining traction for long-range dependency extraction, setting new performance records on several medical benchmarks~\cite{gu2024mamba,wang2024mamba,xing2024segmamba}.

However, a critical drawback accompanies this relentless pursuit of accuracy: model complexity and parameter count escalate sharply. Whether they are U-Net variants embellished with sophisticated modules or modern Transformer- and Mamba-based architectures, the number of trainable parameters easily reaches several tens or even hundreds of millions.

Meanwhile, the conspicuous domain gap between natural and medical images has fuelled a parallel trend, collecting millions of unlabeled images to pre-train domain-specific, self-supervised backbones. Such endeavours have proliferated across X-ray, fundus, pathology, and colonoscopy imaging~\cite{xu2024whole,li2020self,wang2023knowledge,jong2025gastronet}. Typically, a ResNet or Vision Transformer is first pre-trained with contrastive or masked-image modelling, then decapitated and appended with an upsampling decoder to form a segmentation network, before being fine-tuned on public benchmarks. In colonoscopy segmentation, this pipeline consistently outperforms its non-pre-trained counterparts. Yet, the ResNet-50 encoder alone carries 25 M parameters; coupled with a decoder, the overall model balloons to roughly 80 M in our measurements~\cite{jong2025gastronet}.

More recently, promptable segmentation models such as SAM have demonstrated mask-generation capabilities without task-specific annotations, offering a new paradigm for medical image analysis. Nevertheless, their clinical adoption remains limited: the sheer model size hampers deployment, inference is computationally expensive, and their stability and accuracy on dedicated lesion segmentation tasks still lag behind specially trained lightweight networks~\cite{2025Medical}. In particular, the need for per-frame manual prompts from clinicians renders real-time video-stream segmentation clinically impractical.

In summary, although contemporary approaches have matured in accuracy and continue to set new performance records, they share a common deployment bottleneck. The performance gains conferred by ever-larger architectures and self-supervised pre-training are approaching saturation, while the resulting model bulk renders GPU acceleration almost mandatory for real-time inference. This contradiction severely restricts the penetration of state-of-the-art algorithms into resource-constrained environments such as primary hospitals, mobile terminals, and embedded endoscopic systems, and represents the foremost obstacle to translating research breakthroughs into routine clinical practice.

\subsection{Design of lightweight colonoscopy segmentation model}

In the quest for lightweight colonoscopic segmentation, a growing body of work has sought a pragmatic balance between model size and accuracy. Polyp-PVT was the first to import Pyramid Vision Transformer weights pretrained on natural images into polyp segmentation; by leveraging multi-scale global features, it achieved SOTA results in 2022 and firmly established the ViT family in this domain~\cite{dong2021polyp,wang2021pyramid}. Building on this, Polyp-LVT adopted a lightweight ViT backbone, integrated a scale-fusion module and a refined loss, and attained superior segmentation at a smaller footprint~\cite{lin2024polyp}. Another study refined DeepLabV3+ by coupling it with MobileNetV3 and retuning the ASPP dilation rates from 12-24-36 to 8-22-36 to better match polyp morphology, yielding a 6.18 M-parameter model~\cite{jeong2023lightweight}.

Nevertheless, these efforts expose a shared limitation of current lightweight research: compression targets are typically set to the medium scale of around 10M parameters. Although this is already an order of magnitude smaller than the dozens-of-M baselines, it is still insufficient for fluent real-time inference on commodity CPUs, so the bottleneck to ubiquitous deployment remains intact. Colonoscopic or capsule-colonoscopic systems must therefore be paired with GPU servers to realize real-time video-based polyp detection.

An inescapable observation is that, with the advance of self-supervised learning and the collection of large-scale in-domain endoscopic images, simply taking a ResNet or ViT encoder pretrained on massive datasets and appending an upsampling head often outperforms most carefully crafted specialized architectures~\cite{wang2023knowledge,jong2025gastronet}. This compels us to recalibrate the central mission of lightweight research: now that the path to “higher accuracy” is clear, we should instead ask, “When parameters are strictly rationed for CPU-side real-time segmentation, where is the fundamental performance floor?”

Motivated by the above, we recast lightweight polyp-segmentation research into two distinct paradigms:

\begin{itemize}
    \item \textbf{Accuracy-first lightweight}: parameter budget is relaxed (e.g., $>1$ M) and the goal is to approach or surpass the accuracy of large-scale models;
    \item \textbf{Extreme-compression lightweight}: parameters are strictly capped (e.g., $<0.3$ M) to meet real-time, CPU or embedded device only video processing, seeking the optimal accuracy-efficiency trade-off and establishing a performance baseline for severely resource-limited scenarios.
\end{itemize}

Most existing studies follow the first paradigm, leaving the second largely unexplored. To fill this gap, we propose an ultra-lightweight model with fewer than 0.13 M parameters. Rather than competing with architectures an order of magnitude larger, our core contribution is to establish, for the first time, a strong performance baseline under the extreme-compression paradigm: our model achieves markedly leading performance, including competitive accuracy and zero-shot inference capability, while using only 0.13 M parameters. This work demonstrates that high-quality AI can still be deployed in extremely resource-constrained environments and provides a reference for future ultra-lightweight research.

\section{Method}
\label{sec:method}

In medical image segmentation, native ultra-lightweight architectures (<0.1M parameters) have been successfully validated primarily in dermoscopy, where large-scale annotated datasets and well-defined single-lesion ROI structures enable the development of extreme-compression models. Among all medical imaging modalities, dermoscopy and colonoscopy represent the only two domains that simultaneously satisfy three necessary conditions for native ultra-lightweight model development: (i) visible-spectrum surface imaging, (ii) single-region-of-interest segmentation task structure, and (iii) public availability of large-scale annotated benchmarks. Consequently, dermoscopy constitutes the sole mature source of validated sub-0.1M segmentation architectures. We therefore systematically evaluated two representative dermoscopic models, EGE-UNet and LB-UNet~\cite{ruan2023ege,xu2024lb}. Both build on an ultra-slim U-shaped CNN, inject deep supervision, and incorporate Hadamard-product-based linear attention, remaining below 0.1M parameters while delivering strong results on public dermoscopy benchmarks. LB-UNet---equipped with dual-region and boundary supervision and a group-shuffle attention block---exhibits superior boundary awareness and discriminative power on colonoscopic data and was consequently adopted as our initial backbone.

\begin{figure*}[h]
    \centering
    \includegraphics[width=\textwidth]{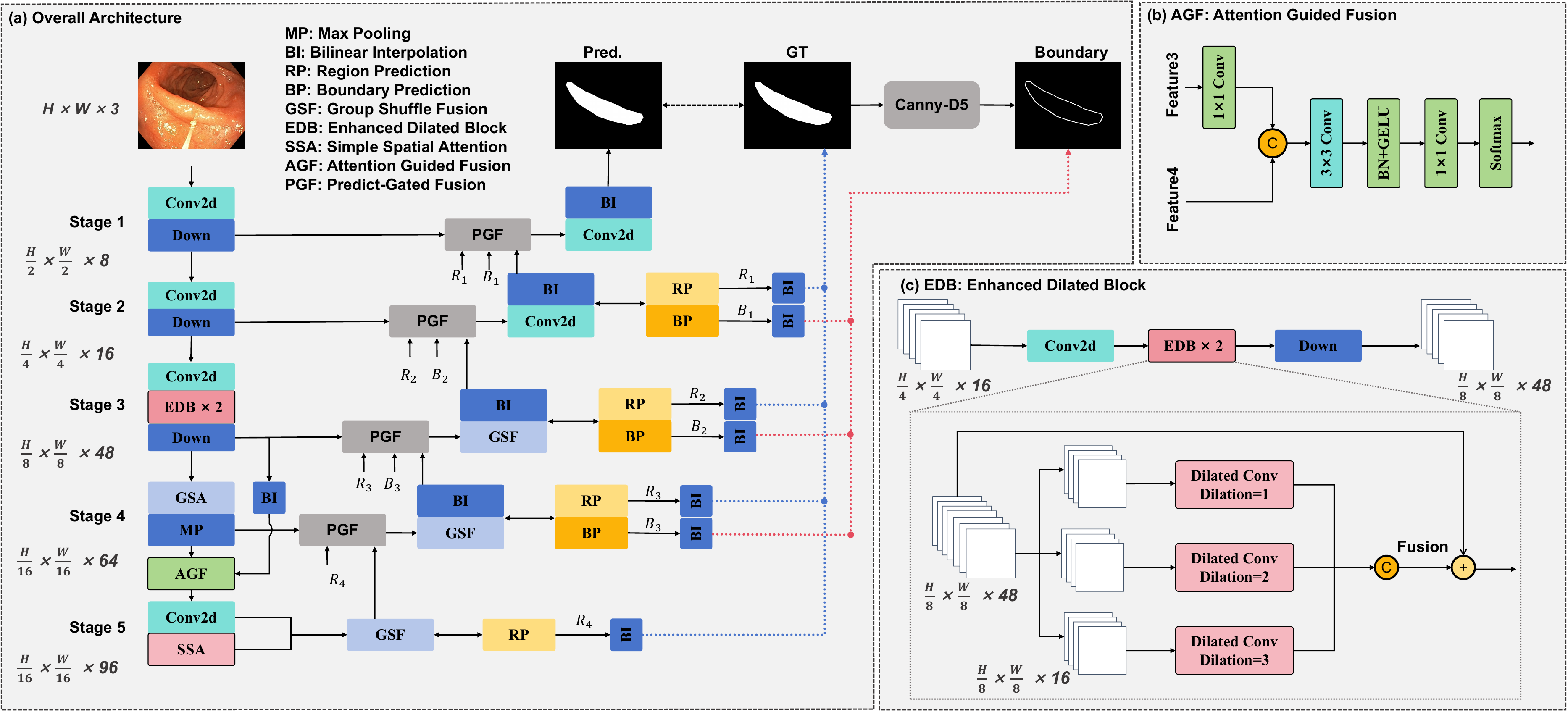} 
    \caption{The architecture of our proposed methods.}
    \label{fig1}
\end{figure*}

Nevertheless, the colonoscopic lumen presents challenges that are qualitatively distinct from dermoscopic skin surfaces: varying illumination modalities, multi-center data aggregation, and indistinct polyp boundaries demand richer contextual representations. The original LB-UNet suffers from a limited receptive field and insufficient cross-center robustness, rendering it ill-suited for these challenges without structural adaptation. To address this, we performed a series of colonoscopy-oriented architectural re-designs on LB-UNet at a restrained parameter cost. The result is an ultra-lightweight segmentation network, UltraSeg-108K, with merely 108K parameters, together with its multi-center, multi-modal extension, UltraSeg-130K. Figure~\ref{fig1} illustrates the overall framework of UltraSeg-130K; the following subsections elaborate on the design rationale and implementation details of each module.

\subsection{Architectural refinements}

LB-UNet was originally tailored for dermoscopy and adopts the channel schedule [8, 16, 24, 32, 48, 64], which keeps many shallow-detail features. Colonoscopic frames, however, contain cluttered backgrounds, highly variable polyp shapes, and blurred boundaries; hence, stronger deep-semantics extraction is required.

We therefore reshape the channel list to [8, 16, 48, 64, 96]. By removing one downsampling stage and simultaneously expanding mid-/high-level channels, we obtain two key benefits: (i) the effective receptive field is enlarged, allowing the network to integrate broader contextual cues to detect polyps that visually merge with the mucosa; (ii) the discriminative power of deep features is strengthened, improving encoding of subtle morphological differences along indistinct polyp borders.

Regarding supervisory signals, LB-UNet iteratively generates boundary ground truth with a genetic algorithm, an expensive and opaque procedure. We replace it with a standard Canny edge detector (5 × 5 structural kernel). The new pipeline shortens boundary-label generation from hours to seconds, is fully deterministic and interpretable, and yields no degradation in segmentation accuracy.

Finally, we redesigned the Predictive Feature-fusion Module. The original version fuses region and boundary predictions through manually tuned, fixed hyperparameters, offering no adaptability. We propose the Predict-Gated Fusion module, whose core innovation is a learnable weight vector coupled with a lightweight feature adaptor. The fusion process is formulated as:

\begin{equation}
\mathbf{x}_{\mathrm{out}} = \mathbf{x}_1' + \mathbf{x}_2 + \alpha \cdot \sigma(\mathbf{P}_{\mathrm{region}}) \odot \mathbf{x}_2 + \beta \cdot \mathbf{P}_{\mathrm{boundary}} \odot \mathbf{x}_2
\label{eq:pgf}
\end{equation}

where $\mathbf{x}_1'$ denotes the decoder feature calibrated by the lightweight adaptor, $\mathbf{x}_2$ is the encoder feature delivered by the skip connection, $\sigma(\cdot)$ the Sigmoid function, $\mathbf{P}_{\mathrm{region}}$ and $\mathbf{P}_{\mathrm{boundary}}$ the region and boundary probability maps, and $\odot$ the Hadamard product. Learnable scalar weights $\alpha$ (region\_weight) and $\beta$ (boundary\_weight) replace the previously hand-tuned coefficients, enabling the fusion strategy to adapt automatically to the data distribution. This not only strengthens the model's capacity to capture the complex appearances of colonoscopic polyps but also eliminates the uncertainty inherent in manual hyper-parameter tuning. 

Moreover, we retain the Group-Shuffle Attention module and the separate region and boundary prediction heads from the original LB-UNet.

\subsection{Enhanced Dilated Block (EDB)}

In the encoder, shallow layers capture local details such as color and texture, whereas deep layers learn highly abstract semantics. The third encoder block (encoder3) sits exactly at the transition from local details to global semantics: its feature map retains sufficient spatial resolution while already possessing rich semantic depth, making it an ideal place to inject multi-scale context.

Building upon these structural refinements, we design the Enhanced Dilated Block (EDB) to enlarge the receptive field without inflating parameters. The input channels are evenly split into three groups, each undergoing depth-wise convolution with dilation rates 1, 2, and 3 in parallel. This allows the layer to simultaneously gather local fine structures, medium-range context, and larger global cues. The three branches are concatenated and fused by a $1\times1$ convolution.

Let the input tensor be $\mathcal{X}\in\mathbb{R}^{C\times H\times W}$. The EDB implements the mapping
\begin{equation}
\mathrm{EDB}(\mathcal{X})=\mathrm{PWConv}\Bigl(\,\|_{i=1,2,3}\;\mathrm{DWConv}_{d=i}(\mathcal{X}_{i};k=3,g=C/3)\Bigr),
\end{equation}

where $\|$ denotes channel-wise concatenation, $\mathcal{X}_{i}$ is the $i$-th equal-split partition of $\mathcal{X}$, $\mathrm{DWConv}_{d=i}$ is depth-wise convolution with dilation rate $i$, and $\mathrm{PWConv}$ is a $1\times 1$ convolution for fusion.

For $C=48$, two cascaded EDBs introduce 50k parameters, yet FLOPs drop by 0.04 G compared with the same-stage vanilla bottleneck (see Table \ref{tab:abl_mixed}), verifying that grouped dilated convolutions enlarge the receptive field without inflating computation.

Two cascaded blocks give a theoretical radius
\begin{equation}
\mathrm{RF}=1+2(k-1)\sum_{i=1}^{3}d_{i}=13\;\text{pixels on the feature map},
\end{equation}

equivalent to $13\times 4=52$ pixels on the input image (encoder3 stride 4), covering a $\sim$50-pixel diameter neighbourhood.

In other words, the network can dynamically integrate multi-scale information within a 50-pixel diameter neighbourhood—mimicking the human ability to reason “from a spot to a patch.” This progressive perception provides the spatial context essential for delineating blurred polyp boundaries and for distinguishing lesions from the complex intestinal background.

Inside each EDB, the 48 input channels are evenly split into 3 branches (16 channels each); every branch undergoes depth-wise dilated convolution, so each filter acts on exactly one channel, imposing structured sparsity and intra-group competition.

\subsection{Cross-Layer Lightweight Fusion (CLLF)}

Improving generalization across multi-center and multi-modal colonoscopy images usually relies on stacking channels or heavy self-attention, both violate our "parametric minimalism" principle. We therefore design \textbf{Cross-Layer Lightweight Fusion (CLLF)}, an ultra-light attention block operating at the encoder-decoder bottleneck. CLLF comprises two functionally coupled stages that share the same $16\!\times\!16$ spatial resolution: (i) Attention Guided Fusion (AGF), and (ii) Simple Spatial Attention (SSA).

\textbf{Stage 1: AGF.} 
Stage-3 features are projected to $64\!\times\!16\!\times\!16$ via $1\!\times\!1$ conv and concatenated with Stage-4 ($128\!\times\!16\!\times\!16$). A $3\!\times\!3$ conv $\rightarrow$ BN $\rightarrow$ GELU $\rightarrow$ $1\!\times\!1$ conv produces two spatial masks $\alpha,\beta$ ($1\!\times\!16\!\times\!16$, $\alpha+\beta=1$ pixel-wise). The fused feature is $\alpha\odot\text{Stage-3}'+\beta\odot\text{Stage-4}'$ ($64\!\times\!16\!\times\!16$).

\textbf{Stage 2: SSA.} 
The fused tensor is averaged along the channel axis and passed through Sigmoid to obtain a $1\!\times\!16\!\times\!16$ attention gate. The feature is then re-weighted with a learnable residual scalar $\gamma$ (initialised to $0.3$):

\begin{equation}
\mathbf{x}=(1-\gamma)\cdot\mathbf{x}_{\mathrm{fused}}+\gamma\cdot\mathbf{x}_{\mathrm{att}}.
\end{equation}

This two-stage design retains global context while suppressing redundant responses; $\gamma$ converges automatically during training. The entire CLLF block consistently boosts segmentation generalisation across centers and modalities, requires no explicit label correction, and adds $<$0.022M parameters, fully satisfying the stringent resource budget of CPU-side real-time inference.

\subsection{Loss function}

To fully exploit deep supervision and strengthen boundary awareness, we adopt a dual-task loss that simultaneously supervises region and edge predictions. The total loss is

\begin{equation}
\mathcal{L}_{\text{total}} = \mathcal{L}_{\text{region}} + \mathcal{L}_{\text{boundary}} + \mathcal{L}_{\text{gt}}
\end{equation}

where

\begin{itemize}
  \item $\mathcal{L}_{\text{region}}$ is the final region-segmentation loss computed as the BceDiceLoss (binary cross-entropy + Dice).
  \item $\mathcal{L}_{\text{boundary}}$ deep-supervises boundary predictions at three decoder levels with weights increasing with depth (0.1, 0.2, 0.3).
  \item $\mathcal{L}_{\text{gt}}$ deep-supervises intermediate region predictions at four decoder levels with weights (0.1, 0.2, 0.3, 0.4).
\end{itemize}

\section{Experiments and results}
\label{sec:experiment}

\subsection{Baseline models and experimental setup}

To systematically evaluate the performance of the proposed model, this study compares it with three categories of representative baseline models:

(1) \textbf{Standard and heavyweight baselines ($>$7M).} This category includes UNet-Base (31M parameters, channel configuration [64,128,256,512,1024]) and UNet-Medium (7.76M parameters, [32,64,128,256,512])~\cite{ronneberger2015u}, both employing vanilla $3\times3$ convolutions. The latter represents the widely adopted third-party implementation in the polyp segmentation community. Additionally, Polyp-PVT~\cite{wang2022pvt} and PraNet~\cite{hu2026pranet}, with ImageNet pre-training, serve as heavyweight references.

(2) \textbf{Lightweight competitors (0.3M--2M).} This group comprises DSConv-based variants derived from UNet-Medium by replacing standard convolutions with depthwise-separable convolutions: UNet-Medium-DS (1.5M, [32,64,128,256,512]), UNet-Light-DS (0.86M, [24,48,96,192,384]), and UNet-Small-DS (0.39M, [16,32,64,128,256]). FastSCNN~\cite{poudel2019fast}, a general-purpose lightweight segmentation network, is also included.

(3) \textbf{Extremely lightweight ($<$0.3M).} This regime features native ultra-light architectures including LB-UNet~\cite{xu2024lb} and EGE-UNet~\cite{ruan2023ege}, which adopt DSConv-based reduced-channel U-Net backbones with task-specific optimizations originally validated in dermatoscopic segmentation. UNet-Tiny-DS (0.1M, [8,16,32,64,128]) represents the most aggressive compression. MobileUNet~\cite{2021Real} is included as a lightweight competitor. Our proposed UltraSeg-108K and UltraSeg-130K operate in this regime, optimized for CPU-real-time polyp segmentation.

A detailed comparison of parameters, FLOPs, and checkpoint sizes is presented in Table~\ref{tab:model-complexity}.

\begin{table*}[h]
\caption{Model complexity comparison.}
\label{tab:model-complexity}
\centering
\begin{tabular}{lccc}
\toprule
\textbf{Models} & \textbf{Params (M)} & \textbf{FLOPs (G)} & \textbf{Checkpoint Size (MB)} \\
\midrule
\multicolumn{4}{l}{\textit{Extremely lightweight ($<$0.3M)}} \\
LB-UNet         & 0.038 & 0.098 & 0.3 \\
EGE-UNet        & 0.053 & 0.072 & 0.4 \\
UNet-Tiny-DS    & 0.102 & 0.290 & 0.5 \\
UltraSeg-108K   & 0.108 & 0.144 & 0.6 \\
UltraSeg-130K   & 0.130 & 0.149 & 0.6 \\
MobileUNet      & 0.140 & 0.170 & 0.6 \\
\midrule
\multicolumn{4}{l}{\textit{Lightweight competitors (0.3M--2M)}} \\
FastSCNN        & 0.367 & 2.210 & 1.5 \\
UNet-Small-DS   & 0.389 & 1.007 & 1.6 \\
UNet-Light-DS   & 0.861 & 2.151 & 3.5 \\
UNet-Medium-DS  & 1.519 & 3.723 & 5.9 \\
\midrule
\multicolumn{4}{l}{\textit{Standard and heavyweight baselines ($>$7M)}} \\
UNet-Medium     & 7.76  & 13.75 & 29.7 \\
Polyp-PVT$^{\dagger}$       & 25.1 & 10.02 & 95.9 \\
PraNet$^{\dagger,\ddagger}$  & 30.8 & 11.20  & 353.1 \\
UNet-Base       & 31.0  & 54.80  & 119.0 \\
\bottomrule
\end{tabular}
\\[3pt]
\footnotesize $^{\dagger}$ ImageNet pre-training required.\\
\footnotesize $^{\ddagger}$ Evaluated via the official PraNet-V2 re-implementation with PVTv2-B2 backbone.
\end{table*}

All experiments were conducted on a single NVIDIA 5090 GPU under PyTorch. Each experiment was run three times with fixed random seeds. The default setting is $256\times256$ resolution, batch size 4, Adam optimizer with learning rate $3\times10^{-4}$, 100 epochs, and early stopping after 10 epochs without validation Dice improvement. For resolution robustness validation, additional experiments at $352\times352$ were conducted under identical protocols. 10\% training data was held out for validation. Detailed configurations are in our GitHub repository and supplementary files.

\subsection{Datasets and evaluation metrics}

The experiments in this study were conducted on four authoritative polyp segmentation datasets: CVC-ClinicDB~\cite{CVC-ClinicDB}, Kvasir-SEG~\cite{jha2020kvasir}, PolypGen~\cite{ali2106polypgen}, and PolypDB~\cite{jha2024polypdb}. CVC-ClinicDB and Kvasir-SEG comprise single-center, single-device data; PolypGen aggregates images from six medical centers; and PolypDB contains five imaging modalities (BLI, FICE, LCI, NBI, WLI) from three distinct centers.

Each dataset was first split 80:20 into training and test sets, then pooled to form a unified mixed training set and mixed test set. All main results in this paper report performance under this protocol. External validation was conducted on two completely unseen datasets, ETIS-Larib~\cite{dumitru2023using} and BKAI-IGH~\cite{son2024polyps}, comprising 196 and 1,000 colonoscopy images with expert-annotated masks, respectively. Models trained on the mixed dataset were directly evaluated on these external sets without fine-tuning.

For comprehensive validation, we additionally report: (i) \textit{per-dataset intra-dataset results} Training and testing on individual datasets showcase a limited samples scenario(CVC, Kvasir, PolypGen, PolypDB); (ii) \textit{instrument segmentation} To assess the model’s general-purpose segmentation capability under colonoscopic imaging conditions, we additionally evaluated it on the Kvasir-Instrument dataset~\cite{kvasirinst}, a surgical-tool segmentation benchmark. In this setting, the task is no longer polyp delineation, but the comparatively simpler detection of procedure instruments. Please refer to the supplementary document for the training evaluation results per dataset.

The naturally imbalanced multi-center and multi-modal distribution of PolypGen and PolypDB persists in both pooled training and test subsets. Such severe skew poses a formidable challenge for generalization, yet renders evaluation more clinically relevant. Table~\ref{table:datafeature} summarizes dataset characteristics.

\begin{table*}[h]
\caption{Overview of datasets in this study}
\label{table:datafeature}
\begin{center}
\resizebox{\textwidth}{!}{
\begin{tabular}{ccccc}  
\toprule
\textbf{Dataset} & \textbf{Data Source} & \textbf{Task} & \textbf{Num. of Images} & \textbf{Usage} \\  
\midrule
CVC & Single-center, single-device & Polyp segmentation & 612 & Development \\
Kvasir & Single-center, single-device & Polyp segmentation & 1,000 & Development \\
Kvasir-Inst & Single-center, single-device & Instrument segmentation & 590 & Development \\
PolypGen & Multi-center (6 institutions) & Polyp segmentation & 1,537 & Development \\
PolypDB & Multi-modalities (5 modalities)  & Polyp segmentation & 3,934 & Development \\
ETIS-Larib& Single-center, single-device & Polyp segmentation & 196 & External validation \\

BKAI-IGH& Single-center, single-device & Polyp segmentation & 1,000 & External validation \\

\bottomrule
\end{tabular}
}
\begin{flushleft}
\footnotesize
\textit{Note:} CVC = CVC-ClinicDB; Kvasir = Kvasir-SEG; Kvasir-Inst = Kvasir-Instrument. 
Datasets marked as "Development" were used for model training and internal validation, 
while "External validation" datasets were used solely for independent testing.
\end{flushleft}
\end{center}
\end{table*}

In this study, the Dice coefficient is employed as the primary evaluation metric for all experiments, with IoU and 95th percentile Hausdorff Distance (HD95, in pixels) additionally reported. The corresponding formulas are given below:

\begin{equation}
    \text{Dice} = \frac{2|X \cap Y|}{|X| + |Y|} = \frac{2\text{TP}}{2\text{TP} + \text{FP} + \text{FN}}
    \label{eq:dice}
\end{equation}

\begin{equation}
    \text{IoU} = \frac{|X \cap Y|}{|X \cup Y|} = \frac{\text{TP}}{\text{TP} + \text{FP} + \text{FN}}
    \label{eq:iou}
\end{equation}

\begin{equation}
    \text{HD95} = P_{95}\left(\left\{\min_{y \in Y}d(x,y)\right\}_{x \in X} \cup \left\{\min_{x \in X}d(x,y)\right\}_{y \in Y}\right)
    \label{eq:hd95}
\end{equation}

where $X$ and $Y$ denote the predicted and ground truth segmentation masks, respectively; TP, FP, FN represent true positives, false positives, and false negatives; $d(\cdot,\cdot)$ denotes the Euclidean distance, and $P_{95}$ denotes the 95th percentile. Notably, Dice and IoU measure region overlap, where higher values indicate better segmentation accuracy. Conversely, HD95 quantifies boundary discrepancy, where lower values signify superior boundary alignment and reduced outlier sensitivity.

\subsection{Main results}

We unified and merged four colonoscopic polyp segmentation datasets, partitioned into training and test sets at an 8:2 ratio. As shown in Table~\ref{tab:mixeddata}, our proposed extremely lightweight solution achieves the best-in-class performance among all lightweight competitors. Specifically, at \textbf{256$\times$256} resolution, UltraSeg-108K attains a Dice score of 0.7884, while UltraSeg-130K reaches 0.8038 with an HD95 of 37.31 pixels. These results not only significantly outperform other lightweight models but also approach the performance of UNet-Medium-DS. Notably, the HD95 metric indicates superior boundary delineation accuracy. Although Polyp-PVT, PraNet, and UNet-Base achieve higher absolute performance, UltraSeg-130K, with merely 0.13M parameters, delivers 96.6\% of the Dice performance of the 7.76M-parameter UNet-Medium. Its HD95 is only marginally inferior to the four standard and heavyweight baselines, demonstrating remarkably sensitive boundary perception given its 0.1M parameter budget. Furthermore, we conducted external validation on two completely unseen datasets, ETIS-Larib~\cite{dumitru2023using} and BKAI-IGH~\cite{son2024polyps}, enabling an unbiased assessment of cross-domain generalization.

As reported in Table~\ref{tab:mixeddata}, UltraSeg-130K achieves 0.6821 Dice on ETIS-Larib and 0.8001 on BKAI-IGH. The performance gaps relative to UNet-Base shrink to merely 2.0 and 1.6 percentage points, respectively. While outperforming all other lightweight models, it even surpasses UNet-Medium in Dice on both external sets. This demonstrates that extremely lightweight polyp segmentation, when supported by principled architectural design, can achieve compelling generalization.

\begin{table*}[h]
\centering
\small
\caption{Mixed-dataset segmentation performance and external validation. Segmentation performance at \textbf{256$\times$256} resolution.}
\label{tab:mixeddata}
\resizebox{\textwidth}{!}{%
\begin{tabular}{l c ccc ccc ccc}
\toprule
\multirow{2}{*}{Model} &
\multirow{2}{*}{Params (M)} &
\multicolumn{3}{c}{Mixed Dataset} &
\multicolumn{3}{c}{ETIS-Larib External} &
\multicolumn{3}{c}{BKAI-IGH External} \\
\cmidrule(lr){3-5}\cmidrule(lr){6-8}\cmidrule(lr){9-11}
& & Dice$\uparrow$ & IoU$\uparrow$ & HD95$\downarrow$
& Dice$\uparrow$ & IoU$\uparrow$ & HD95$\downarrow$
& Dice$\uparrow$ & IoU$\uparrow$ & HD95$\downarrow$ \\
\midrule
\multicolumn{11}{l}{\textit{Standard and heavyweight baselines ($>$7M)}} \\
UNet-Base       & 31.0  & 0.8478 & 0.7412 & 33.48 & 0.7094 & 0.5830 & 38.09 & 0.8163 & 0.7045 & 26.28 \\

PraNet\dag   & 30.8&0.8899&	0.8139&23.04&0.8280&0.7261&35.35&0.8449&0.7419&	21.41
 \\
Polyp-PVT\dag   & 25.11 & 0.8976 & 0.8324 & 20.27 & 0.8221 & 0.7223 & 27.80 & 0.8492 & 0.7479 & 19.80 \\

UNet-Medium          & 7.76  & 0.8318 & 0.7240 & 36.14 & 0.6797 & 0.5460 & 45.71 & 0.7958 & 0.6768 & 32.86 \\
\midrule
\multicolumn{11}{l}{\textit{Lightweight competitors (0.3M--2M)}} \\
UNet-M-DS       & 1.519 & 0.8131 & 0.7010 & 40.82 & 0.6295 & 0.4908 & 47.59 & 0.7764 & 0.6539 & 33.06 \\
UNet-L-DS       & 0.861 & 0.8034 & 0.6816 & 46.30 & 0.6234 & 0.4843 & 58.69 & 0.7697 & 0.6467 & 37.12 \\
UNet-S-DS       & 0.389 & 0.7704 & 0.6570 & 44.23 & 0.5826 & 0.4484 & 61.01 & 0.7337 & 0.6021 & 36.52 \\
FastSCNN        & 0.367 & 0.6685 & 0.5182 & 80.70 & 0.4686 & 0.3245 & 100.94 & 0.6209 & 0.4701 & 78.80 \\

\midrule
\multicolumn{11}{l}{\textit{Extremely lightweight ($<$0.3M)}} \\
MobileUNet      & 0.140 & 0.4879 & 0.3229 & 93.49 & 0.2248 & 0.1346 & 102.39 & 0.4443 & 0.3013 & 94.29 \\
UNet-T-DS       & 0.102 & 0.6716 & 0.5201 & 50.24 & 0.4602 & 0.3375 & 67.61 & 0.6097 & 0.4667 & — \\
EGE-UNet        & 0.053 & 0.7079 & 0.5632 & 59.22 & 0.5547 & 0.4118 & 66.34 & 0.6878 & 0.5438 & 51.04 \\
LB-UNet         & 0.038 & 0.7685 & 0.6425 & 40.74 & 0.6437 & 0.5062 & 57.12 & 0.7700 & 0.6435 & 35.15 \\
\textbf{UltraSeg-108K} & \textbf{0.108} & \textbf{0.7884} & \textbf{0.6676} & \textbf{39.96} & \textbf{0.6736} & \textbf{0.5387} & \textbf{53.66} & \textbf{0.7873} & \textbf{0.6662} & \textbf{31.97} \\
\textbf{UltraSeg-130K} & \textbf{0.130} & \textbf{0.8038} & \textbf{0.6882} & \textbf{37.31} & \textbf{0.6821} & \textbf{0.5478} & \textbf{50.24} & \textbf{0.8001} & \textbf{0.6796} & \textbf{30.50} \\
\bottomrule
\multicolumn{11}{l}{\footnotesize —: HD95 omitted due to excessive calculation error. \dag~denotes methods requiring ImageNet pre-training. }
\end{tabular}}
\end{table*}

Table~\ref{tab:resolution352} presents the performance of key models at \textbf{352$\times$352} resolution. Here, UNet-T-DS denotes the vanilla UNet baseline with DSConv at a comparable parameter scale but without architectural optimization. UNet-Medium and UNet-M-DS represent medium-capacity UNet variants using standard convolution and DSConv, respectively, under identical channel configurations. When trained and evaluated at higher resolution, both UltraSeg variants exhibit consistent performance gains. Although UltraSeg-130K still trails UNet-Medium on the mixed dataset, it surpasses the 1.519M-parameter UNet-M-DS. More importantly, on both external validation datasets, UltraSeg-130K exceeds or closely approaches UNet-Medium in Dice.

These results collectively confirm that our architectural improvements deliver substantial performance advantages over all similarly sized models at both 256 and 352 resolutions, while exhibiting strong zero-shot cross-dataset generalization.

\begin{table*}[h]
\caption{Segmentation performance at \textbf{352$\times$352} resolution. All results are averaged over three fixed random seeds.}
\label{tab:resolution352}
\centering
\small
\resizebox{\textwidth}{!}{%

\begin{tabular}{lcccccccccc}  
\toprule
\multirow{2}{*}{Model} &
\multirow{2}{*}{Params (M)} &
\multicolumn{3}{c}{Mixed Dataset} &
\multicolumn{3}{c}{ETIS-Larib External} &
\multicolumn{3}{c}{BKAI-IGH External} \\
\cmidrule(lr){3-5}\cmidrule(lr){6-8}\cmidrule(lr){9-11}
& & Dice$\uparrow$ & IoU$\uparrow$ & HD95$\downarrow$
& Dice$\uparrow$ & IoU$\uparrow$ & HD95$\downarrow$
& Dice$\uparrow$ & IoU$\uparrow$ & HD95$\downarrow$ \\
\midrule
UNet-Medium & 7.76  & 0.8345 & 0.7311 & 49.47 & 0.6924 & 0.5632 & 58.13 & 0.8120 & 0.6977 & 41.56 \\
UNet-M-DS       & 1.519 & 0.8108 & 0.6958 & 63.98 & 0.6306 & 0.4918 & 72.09 & 0.7846 & 0.6628 & 49.29 \\
LB-UNet         & 0.038 & 0.7726 & 0.6476 & 59.02 & 0.6385 & 0.5012 & 87.18 & 0.7732 & 0.6469 & 56.07 \\
UNet-T-DS       & 0.102 & 0.6711 & 0.5241 & 78.01 & 0.4798 & 0.3482 & 110.66& 0.6366 & 0.4923 & 64.28 \\
\midrule
\textbf{UltraSeg-108K} & \textbf{0.108} & \textbf{0.8060} & \textbf{0.6873} & \textbf{60.12} & \textbf{0.6916} & \textbf{0.5516} & \textbf{75.77} & \textbf{0.7998} & \textbf{0.6797} & \textbf{47.48} \\
\textbf{UltraSeg-130K} & \textbf{0.130} & \textbf{0.8134} & \textbf{0.6989} & \textbf{54.77} & \textbf{0.6960} & \textbf{0.5594} & \textbf{72.40} & \textbf{0.8013} & \textbf{0.6832} & \textbf{44.27} \\
\bottomrule
\end{tabular}}
\end{table*}

\subsection{Inference efficiency comparison on commodity CPUs}

Table~\ref{tab:fps} reports the average frames-per-second (FPS) measured by continuously inferring 1,000 images on CPUs. The Dice column summarizes the Dice scores of different methods on the mixed dataset at 256$\times$256 resolution. The FPS metrics were obtained via single-core inference on an Intel i5-14600K CPU, emulating resource-constrained edge deployment scenarios.

As expected, increasing the inference resolution from 256 to 352 substantially degrades the FPS. Nevertheless, even under strict single-core constraints, all our models consistently exceed 30 FPS. By contrast, larger competitors such as UNet-M-DS are entirely incapable of real-time inference on an i5-14600K CPU, equivalent consumer-grade processors, or typical edge devices. Additional benchmarking results on higher-end devices (i9-14900K series CPUs) are provided in the supplementary materials, and the FPS evaluation code is publicly available on GitHub.

\begin{table*}[h]
\caption{CPU inference speed on Intel i5-14600K (single-core). Dice scores from mixed-dataset training at 256$\times$256.}
\label{tab:fps}
\centering
\small
\begin{tabular}{lcccc}
\toprule
\textbf{Models} & \textbf{Params (M)} & \textbf{Dice$\uparrow$} & \textbf{FPS@256} & \textbf{FPS@352} \\
\midrule

UNet-Base       & 31.0  & 0.8478 & 1.3   & 0.7   \\
PraNet   & 30.8 & 0.8899 &6.0&3.2\\
Polyp-PVT   & 25.11 & 0.8976 & 4.3   & 3.7   \\
UNet-Medium     & 7.76  & 0.8318 & 2.8   & 2.3   \\
UNet-M-DS       & 1.519 & 0.8131 & 5.3   & 4.7   \\
UNet-L-DS       & 0.861 & 0.8034 & 8.4   & 5.1   \\
UNet-S-DS       & 0.389 & 0.7704 & 14.7  & 8.8   \\
FastSCNN        & 0.367 & 0.6685 & 33.6  & 17.2   \\
MobileUNet      & 0.140 & 0.4879 & 276.1 & 105.0 \\
UNet-T-DS       & 0.102 & 0.6716 & 55.6  & 37.7  \\
EGE-UNet        & 0.053 & 0.7079 & 85.1  & 30.6  \\
LB-UNet         & 0.038 & 0.7685 & 53.1  & 54.0  \\
\midrule
\underline{\textbf{UltraSeg-108K}} & \underline{\textbf{0.108}} & \underline{\textbf{0.7884}} & \underline{\textbf{52.2}} & \underline{\textbf{30.8}} \\
\underline{\textbf{UltraSeg-130K}} & \underline{\textbf{0.130}} & \underline{\textbf{0.8038}} & \underline{\textbf{51.7}} & \underline{\textbf{30.3}} \\
\bottomrule
\end{tabular}
\end{table*}

\begin{figure}[h]
\centering
\begin{minipage}{0.48\linewidth}
  \includegraphics[width=\linewidth]{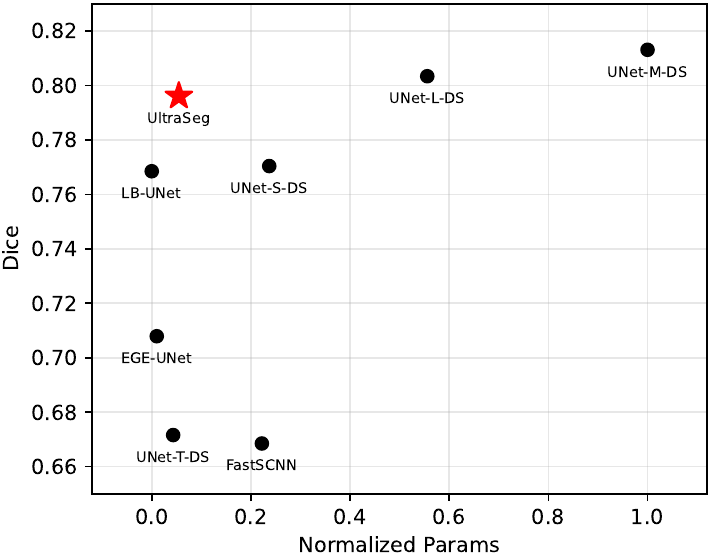}
\end{minipage}\hfill
\begin{minipage}{0.48\linewidth}
  \includegraphics[width=\linewidth]{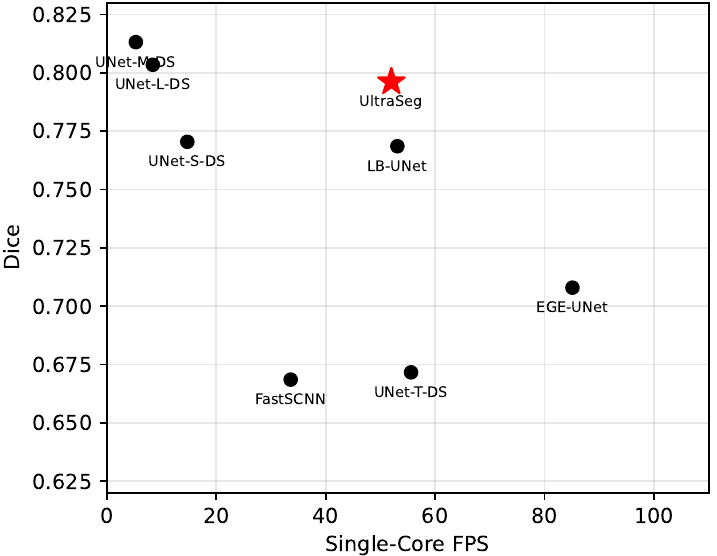}
\end{minipage}
\caption{Left: Params-Dice trade-off; Right: FPS-Dice trade-off on a single-core Intel i5 processor at $256\times256$ resolution.}
\label{fig:sweet_spot}
\end{figure}

Figure~\ref{fig:sweet_spot} presents scatter plots of parameter count \textit{vs.}~Dice (left) and single-core CPU-FPS \textit{vs.}~Dice (right) for lightweight comparison models. Red stars denote our proposed UltraSeg family. In the left panel, UltraSeg-130K achieves 0.8038 Dice with merely 0.13M parameters, matching the accuracy of UNet-L-DS (0.861M parameters, 0.8034 Dice) while exhibiting substantially superior HD95 and markedly stronger zero-shot generalization performance---even surpassing UNet-Medium and outperforming all dedicated lightweight models below 0.3M parameters. In the right panel, the UltraSeg family consistently delivers over 50~FPS inference speed on a single-core Intel i5 processor at 256$\times$256 resolution. The UltraSeg family precisely anchors the critical threshold for CPU real-time inference: simultaneously attaining the optimal sweet spot across parameter efficiency, inference speed, and segmentation accuracy. This redefines "lightweight" from mere parameter compression to "clinically deployable real-time precision", providing the first feasible solution with zero-shot generalization capability for resource-constrained clinical scenarios.

\subsection{Visualization and qualitative analysis}
\label{bnexplain}

\begin{figure*}[h]
    \centering
    \includegraphics[width=\textwidth]{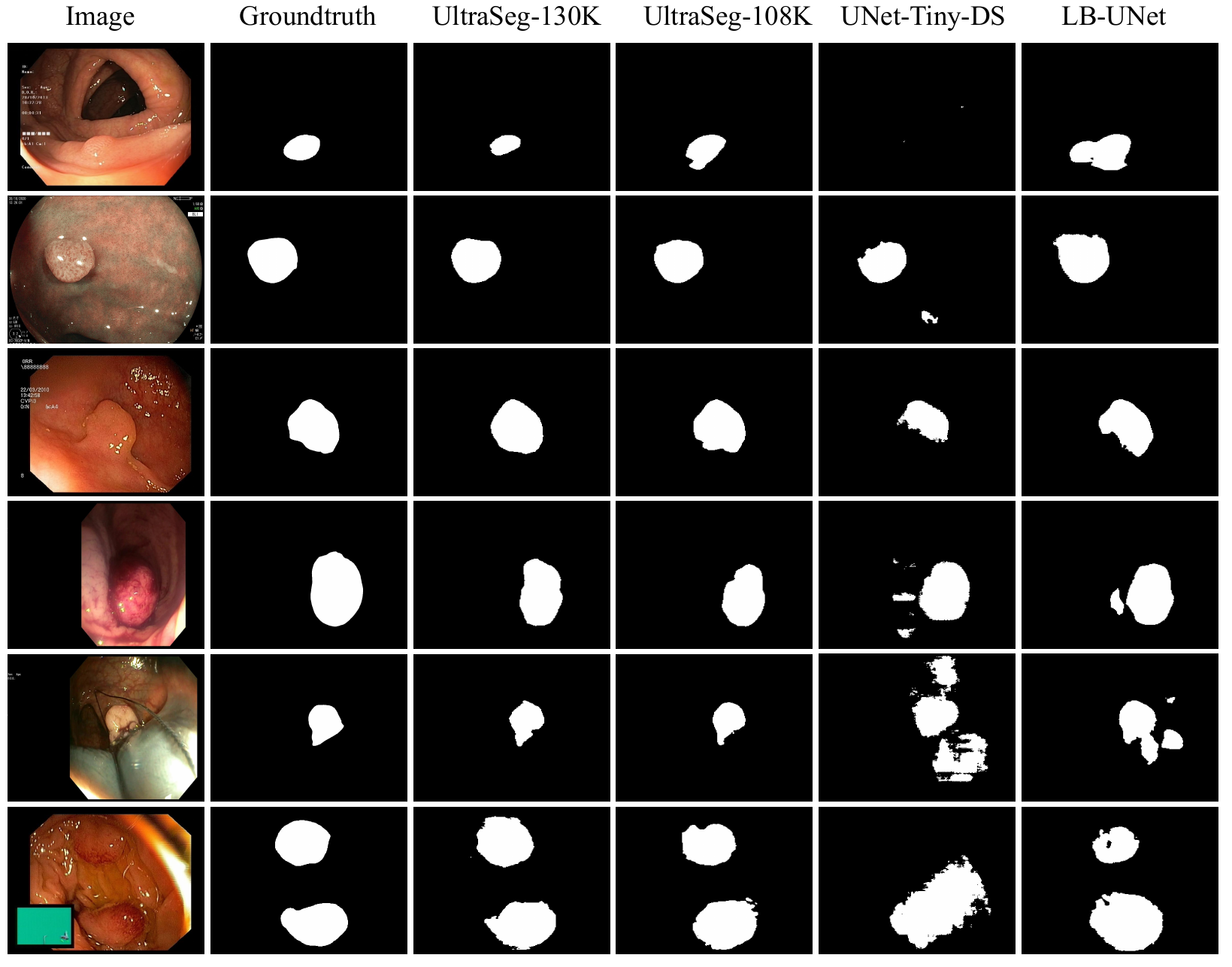} 
    \caption{Qualitative comparisons on lightweight models.}
    \label{img:visual}
\end{figure*}

Figure~\ref{img:visual} compares segmentation results on representative test images among UNet-Tiny-DS, LB-UNet, and our proposed UltraSeg-108K/130K. Regardless of polyp size or quantity, our proposed models consistently surpass competitors owing to fine-grained architectural refinements under an extreme parameter budget. While boundary delineation remains challenging for all models operating under severe parameter constraints, UltraSeg-130K demonstrates visibly improved edge continuity compared to baseline lightweight architectures. The remaining boundary irregularities reflect the fundamental trade-off between model capacity and spatial resolution reconstruction; however, the progressive perception design of EDB partially alleviates this by preserving multi-scale contextual cues, enabling UltraSeg-130K to deliver substantially superior delineation within the constrained budget. Additional qualitative comparisons on challenging zero-shot cases are provided in the Supplementary Material to offer a more comprehensive performance assessment.

\subsection{Cross-Layer Lightweight Fusion gains in per-dataset training}

To further validate the effectiveness of our proposed Cross-Layer Lightweight Fusion in handling heterogeneous, multi-center, multi-imaging-modality data, we conducted independent training and validation experiments on each individual dataset, with detailed results provided in the supplementary materials.

The findings reveal that when restricted to single-source datasets such as Kvasir or CVC, the upgrade from UltraSeg-108K to UltraSeg-130K yields negligible performance gains, indicating that the additional cross-layer fusion capacity is not fully leveraged under homogeneous imaging conditions. Conversely, on multi-center datasets such as PolypDB and PolypGen, UltraSeg-130K exhibits substantial performance improvements, confirming that Cross-Layer Lightweight Fusion effectively captures complementary representations across diverse clinical sources and imaging protocols.

\begin{figure*}[h]
    \centering
    \includegraphics[width=\textwidth]{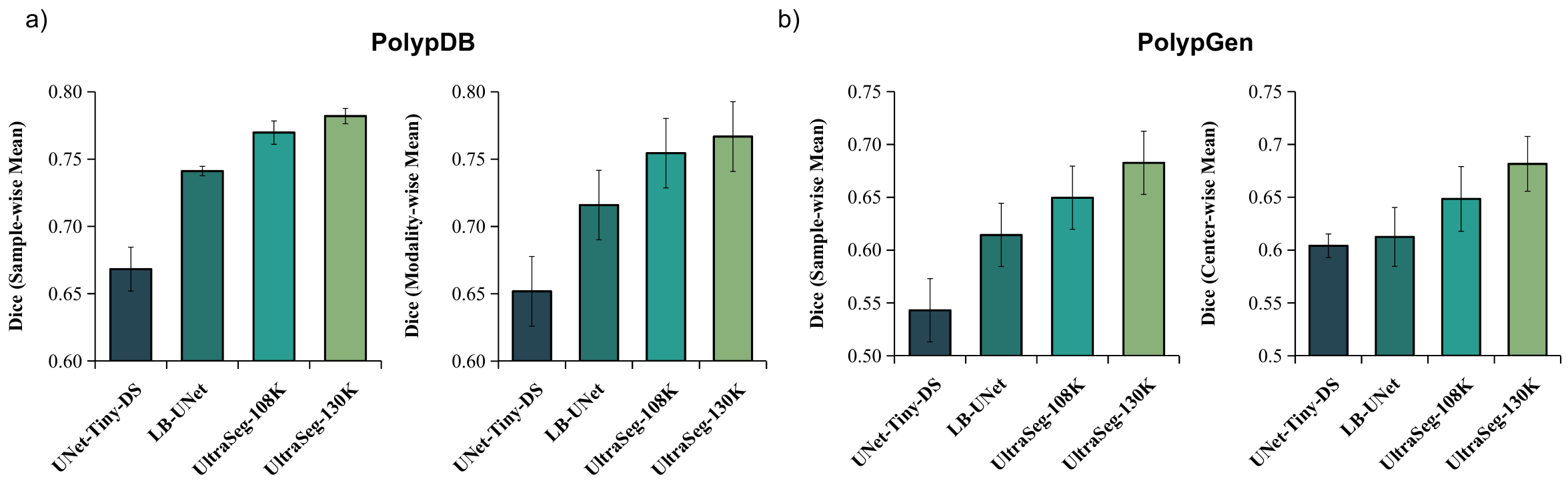} 
    \caption{The figure presents the per-sample mean Dice and the center- or modality-level mean Dice for different models on the PolypDB and PolypGen datasets.}
    \label{img:visualchart}
\end{figure*}

Figure~\ref{img:visualchart} compares the proposed method with the baselines UNet-Tiny-DS and LB-UNet on PolypDB and PolypGen datasets. Building upon UltraSeg-108K, UltraSeg-130K introduces attention-guided skip connections after the 3rd and 4th downsampling stages and adds an extra spatial-attention residual block after the bottleneck.

Experiments show consistent gains in average sample-level Dice and modality-averaged Dice. Our cross-layer attention fusion at the end of downsampling boosts the generalization of a lightweight model during multi-modal, multi-center training, without extra labels and with almost negligible extra parameters. The approach remains robust even when sample sizes are highly imbalanced across centers or modalities.

In summary, this strategy effectively mitigates distribution shift with only a small increase in parameters, ensuring stable segmentation performance across institutions and devices and thus facilitating clinical deployment.

\subsection{Capacity scaling of UltraSeg}
\label{parameterscaling}

Although the preceding results have demonstrated that UltraSeg maintains remarkably strong performance under small channel configurations with merely 0.1M parameters, an unresolved question remains: whether such performance exhibits a scaling effect as channel dimensions increase. To address this, we conducted the following experiments by progressively increasing channel capacities.

We systematically expanded the channel dimensions of UltraSeg to construct three variants: UltraSeg-500K (channels: $[16, 32, 96, 128, 192]$), UltraSeg-1.11M (channels: $[24, 48, 144, 192, 288]$) and UltraSeg-4.38M(channels: $[48,96,288,384,576]$), scaled from the original configuration $[8, 16, 48, 64, 96]$ (130K parameters). All models were trained on the Mixed Dataset using identical training protocols, with results averaged over three random seeds. We compared these against the Transformer-based colonoscopy-specific model Polyp-PVT, UNet-Base, and UNet-Medium.

As shown in Table~\ref{tab:capacity_scaling}, we adopt Dice/Params (Dice score per million parameters at 256 resolution) as the architectural efficiency metric. UltraSeg-130K demonstrates a pronounced advantage with a value of 6.183, surpassing all compared models. This indicates that, under extreme compression conditions with the prerequisite of real-time colonoscopic polyp segmentation on CPU device, the UltraSeg-130K configuration strikes a Pareto-optimal balance between accuracy and deployment feasibility, validating the effectiveness of the proposed architectural design methodology.

As channel dimensions increase, inference speed on a single-core CPU drops sharply: UltraSeg-130K achieves 51.7 FPS on a single-core Intel i5-14600K, satisfying real-time video processing requirements ($>$30 FPS). In comparison, UltraSeg-4.38M falls to merely 7.3 FPS. Furthermore, although Polyp-PVT achieves the highest Dice of 0.8976, its inference speed of 4.3 FPS (requiring GPU acceleration) renders it unsuitable for resource-constrained deployment.

Beyond the real-time optimal 130K configuration, we further investigate whether the architectural advantages of UltraSeg persist across a broader parameter spectrum. Surprisingly, UltraSeg-4.38M achieves a Dice of 0.8503 with only 4.38M parameters, surpassing both UNet-Medium and UNet-Base. This demonstrates that our bottom-up design principles, with Enhanced Dilated Block and Cross-Layer Lightweight Fusion, yield a more efficient backbone than standard U-Net architectures across all parameter regimes. The consistent efficiency spanning from 130K to 4.38M parameters validates that UltraSeg represents a scalable architecture family, offering flexible accuracy--speed trade-offs for diverse clinical scenarios.

\begin{table*}[h]
  \centering
  \caption{Capacity scaling of UltraSeg and comparison with heavyweight baselines. All results are averaged over three fixed random seeds.}
  \label{tab:capacity_scaling}
  \resizebox{\textwidth}{!}{
  \begin{tabular}{@{}lcccccccc@{}}
    \toprule
    Method & Params (M) & FLOPs (G) & FPS@256 (CPU) &  Dice@256 & HD95@256 & Dice vs UNet-B & Efficiency$^*$ \\
    \midrule
    \textbf{UltraSeg-130K} & 0.13 & 0.15 & \textbf{51.7} & 0.8038 &37.31 & 94.8\% & \textbf{6.183} \\
    UltraSeg-500K & 0.50 & 0.53 & 27.8 & 0.8237	&35.36 & 97.1\% & 1.647 \\
    UltraSeg-1.11M & 1.11 & 1.15 & 17.7 & 0.8327&33.86& 98.2\% & 0.749 \\
   UltraSeg-4.38M & 4.38 & 4.44 & 7.3 & \textbf{0.8503}& 32.56& \textbf{100.2\%} & 0.194 \\
    \midrule
    UNet-Medium & 7.76 & 13.75 & 2.8 & 0.8318&36.14 & 98.1\% &0.106 \\

    UNet-Base & 31.04 & 54.8 & 1.3 & 0.8478&33.48 & 100\% & 0.027 \\
    Polyp-PVT$^\dagger$ & 25.11 & 10.02 & 4.3 & 0.8976&20.27 & 105.8\% & 0.035 \\
    \bottomrule
  \end{tabular}}
  \footnotesize{$^*$ Efficiency = Dice / Params (per million). Higher is better.}\\
  \footnotesize{$^\dagger$ Pretrained on ImageNet.}
\end{table*}

\subsection{Ablation Experiments}
\label{ablation}

Table~\ref{tab:abl_mixed} systematically presents the progressive ablation of our proposed framework on the aggregated dataset at resolutions of 256 and 352, alongside comparisons with the lightweight baseline UNet-Tiny-DS and the stronger LB-UNet model. All metrics are averaged across three independent runs to ensure statistical robustness.

First, LB-UNet significantly improves upon UNet-Tiny-DS by introducing joint deep supervision over both regions and boundaries. Building upon this foundation, we incorporate task-specific architectural refinements, denoted as UltraSeg (ST). Despite increasing the parameter count from 38K (LB-UNet) to 60K, marked Dice improvements are observed at both resolutions. The proposed adaptive strategy more effectively captures the distinctive characteristics of colonoscopic images, while streamlining hyperparameter tuning and alleviating reliance on complex preprocessing techniques such as genetic algorithms.

Subsequently, augmenting the third downsampling stage with two consecutive Enhanced Dilated Blocks yields the UltraSeg (ST+EDB) variant. With the parameter count rising to 108K, substantial performance gains are achieved alongside improved boundary delineation accuracy as quantified by HD95. This constitutes our UltraSeg-108K configuration, designed for single-center deployment.

Finally, the CLLF module, specifically tailored for multi-center and multi-modal data, produces UltraSeg-130K, which achieves pronounced performance gains at both resolutions. Throughout the ablation process, fixed channel widths are maintained to strictly preserve the compact model size, thereby enabling real-time inference under resource-constrained conditions. This bottom-up lightweight construction paradigm ensures that incremental parameter growth consistently translates into measurable performance improvements, whereas FLOPs do not increase significantly and may even decrease throughout the progression.

\begin{table*}[h]
\centering
\caption{Ablation study on architectural components under mixed-dataset training. 
         Evaluated at 256$\times$256 and 352$\times$352 resolutions.}
\label{tab:abl_mixed}
\resizebox{0.85\textwidth}{!}{%
\begin{tabular}{lcccccc}
\toprule
\multirow{2}{*}{\textbf{Method}} & 
\multirow{2}{*}{\textbf{Params (M)}} & 
\multirow{2}{*}{\textbf{FLOPs (G)}} & 
\multicolumn{2}{c}{\textbf{256$\times$256}} & 
\multicolumn{2}{c}{\textbf{352$\times$352}} \\
\cmidrule(lr){4-5}\cmidrule(lr){6-7}
& & & Dice$\uparrow$ & HD95$\downarrow$ & Dice$\uparrow$ & HD95$\downarrow$ \\
\midrule
UNet-Tiny-DS\textsuperscript{a} & 0.102 & 0.290 & 0.6716 & 50.24 & 0.6711 & 78.01 \\
LB-UNet\textsuperscript{b} & 0.038 & 0.098 & 0.7685 & 40.74 & 0.7726 & 59.02 \\
\midrule
UltraSeg (ST)\textsuperscript{c} & 0.060 & 0.185 & 0.7726 & 42.46 & 0.7832 & 61.15 \\
UltraSeg (ST+EDB)\textsuperscript{d} & 0.108 & 0.144 & 0.7884 & 39.96 & 0.8060 & 60.12 \\
\underline{\textbf{UltraSeg (ST+EDB+CLLF)}}\textsuperscript{e} & \underline{\textbf{0.130}} & \underline{\textbf{0.149}} & \underline{\textbf{0.8038}} & \underline{\textbf{37.31}} & \underline{\textbf{0.8134}} & \underline{\textbf{54.77}} \\
\bottomrule
\end{tabular}%
}
\begin{flushleft}
\footnotesize
\textsuperscript{a} Vanilla lightweight baseline (using DSConv).\\
\textsuperscript{b} Strong lightweight baseline from dermoscopy.\\
\textsuperscript{c} \textbf{S}tructural \textbf{T}weaks: channel reshape, Canny boundary, Predict-Gated Fusion.\\
\textsuperscript{d} (c) + \textbf{E}nhanced \textbf{D}ilated \textbf{B}lock (2$\times$ at encoder stage 3).\\
\textsuperscript{e} (d) + CLLF (\textbf{A}ttention-\textbf{G}uided \textbf{F}usion + \textbf{S}imple \textbf{S}patial \textbf{A}ttention).\\
\end{flushleft}
\end{table*}

Table~\ref{tab:abl_multiscale} presents the ablation results across varying channel configurations (500K, 1.11M, and 4.38M) at a resolution of 256$\times$256. The experimental results demonstrate that the key architectural refinements proposed in this work yield substantial performance gains for the overall system, and such gains remain robust even under expanded channel configurations.

Notably, in the absence of both EDB and CLLF, merely increasing channel dimensions (from 0.226M to 1.93M parameters) yields no monotonic improvement in either Dice or HD95; in some cases, performance even degrades with larger capacity. This result provides strong evidence that the native performance of our method is not an artifact of channel scaling. Rather, it is the proposed EDB and CLLF modules, operating within an extremely tight parameter budget, that deliver substantial, scalable gains across all capacity regimes. The baseline LB-UNet architecture, even when scaled up, cannot match this efficiency, confirming that our architectural innovations constitute the primary driver of the reported accuracy.

By contrast, the EDB module consistently unlocks significant accuracy improvements across all scales (Dice gains of approximately 5.1\%--8.4\%), while simultaneously reducing computational overhead (FLOPs decrease by 22\%--25\%). Furthermore, the complete CLLF block further refines boundary localization, achieving additional HD95 reductions across all evaluated capacities.

\begin{table*}[h]
\centering
\caption{Module contribution across varying channel configurations (256$\times$256, mixed-dataset training). All experiments were conducted using three fixed random seeds, and the reported results are averaged across these runs.}
\label{tab:abl_multiscale}
\small
\setlength{\tabcolsep}{5pt}
\begin{tabular}{llcccc}
\toprule
\textbf{Scale} & \textbf{Configuration} & \textbf{Params (M)} & \textbf{FLOPs (G)} & \textbf{Dice$\uparrow$} & \textbf{HD95$\downarrow$} \\
\midrule
\multirow{3}{*}{500K} 
 & w/o EDB, w/o CLLF & 0.226 & 0.67 & 0.7605 & 40.70 \\
 & w/ EDB, w/o CLLF & 0.413 & 0.51 & 0.8118 & 38.50 \\
 & \textbf{UltraSeg-500K (full)} & \textbf{0.500} & \textbf{0.53} & \textbf{0.8237} & \textbf{35.36} \\
\midrule
\multirow{3}{*}{1.11M} 
 & w/o EDB, w/o CLLF & 0.496 & 1.47 & 0.7540 & 42.75 \\
 & w/ EDB, w/o CLLF & 0.917 & 1.10 & 0.8297 & 35.13 \\
 & \textbf{UltraSeg-1.11M (full)} & \textbf{1.110} & \textbf{1.15} & \textbf{0.8327} & \textbf{33.86} \\
\midrule
\multirow{3}{*}{4.38M} 
 & w/o EDB, w/o CLLF & 1.930 & 5.69 & 0.7592 & 42.52 \\
 & w/ EDB, w/o CLLF & 3.610 & 4.24 & 0.8432 & 32.63 \\
 & \textbf{UltraSeg-4.38M (full)} & \textbf{4.380} & \textbf{4.44} & \textbf{0.8503} & \textbf{32.56} \\
\bottomrule
\end{tabular}
\end{table*}

\section{Discussion}
\label{sec:discussion}

This study introduces the UltraSeg family, built through progressive architectural refinements under strict parameter budgets. The baseline variant, UltraSeg‑108K (0.108M parameters), establishes a strong foundation for extreme‑compression polyp segmentation. Adding a Cross‑Layer Lightweight Fusion (CLLF) module yields UltraSeg‑130K (0.130M parameters), which delivers substantially improved generalization across multi‑center and multi‑modal datasets with negligible parameter overhead. Following a “minimal‑parameter” principle, every additional module is designed to yield maximum performance gains at the lowest possible parametric cost. The models achieve competitive results on seven public colonoscopy segmentation datasets, convincingly demonstrating that reliable pixel‑level segmentation is attainable under extreme parameter constraints.

Currently, most lightweight polyp segmentation studies follow the "accuracy first" paradigm, with parameter budgets typically around 10M, and generally rely on pre-trained backbone networks on large-scale natural image datasets such as ImageNet\cite{wang2022pvt,lin2024polyp,deng2009imagenet}. Although this paradigm has achieved excellent accuracy in computationally resource-rich environments, its model size and pre-training dependencies make it impossible to achieve real-time inference in pure CPU, resource-limited clinical scenarios. This study aims to explore a completely different paradigm of "extreme compression", with the core constraint of strictly limiting the number of parameters to below 0.3M, training from scratch, and meeting real-time video processing requirements of over 30 FPS on a single CPU core. While our architectural principles are scalable, as evidenced by UltraSeg‑4.38M surpassing U‑Net‑Base accuracy, variants exceeding 0.3M parameters no longer satisfy this real‑time CPU constraint. UltraSeg‑4.38M is presented solely to demonstrate architectural scalability; for the targeted clinical use case of single‑core CPU inference, the 0.13M variant remains the recommended choice. Under this paradigm, traditional lightweight methods based on pre-trained large models cannot be shortlisted due to size and computational overhead. Therefore, the baseline for this work should be a model that is in the same "arena" as ours: a network with fewer than 0.3M parameters, also not relying on external pre-training, and designed for CPU deployment (such as LB-UNet, MobileUNet, and FastSCNN). Our contribution lies in establishing a powerful performance benchmark for the first time in this previously unexplored field of "extreme compression".

This study proposes a lightweight path that contrasts with the mainstream "top-down" paradigm. Most high-accuracy methods rely heavily on visual extractors pre-trained on large natural images, which use attention and multi-scale fusion to achieve high accuracy. However, this approach leads to a significant increase in parameter count and strong GPU dependence, which hinders its clinical implementation. We take advantage of the commonality between dermatoscopy and colonoscopy in terms of surface optical imaging, blurred lesion boundaries, and variable morphology. Starting from the validated ultra-light dermatoscopy skeleton, we perform rigorous structural re-engineering for colonoscopy-specific visual characteristics to obtain an efficient segmentation model with 108K/130K parameters and CPU deployment, constituting a novel architectural paradigm. The core advantage of this process is that it does not passively prune redundant models but actively introduces efficient inductive biases that have been verified in similar visual tasks, thereby achieving steady performance improvements from an extremely low parameter starting point. The lightweight pathway revealed in this study stands in sharp contrast to the prevailing paradigm of “large model + complex modules”. The latter aims to push the performance ceiling in compute-rich environments, whereas our work strives for practical breakthroughs under strictly limited resources. The two approaches are not ranked; they serve distinct clinical goals.

Compared with existing lightweight models whose parameters commonly exceed 1M, this study breaks new ground by pushing model complexity down an entire order of magnitude. Under fair and strictly enforced dataset splits, and constrained to real-time inference of >30 FPS on a single CPU core, our model delivers the best performance reported to date, even though its absolute accuracy still trails large-scale models. This usability leap carries greater practical weight for rolling out AI colonoscopy screening in primary-level or resource-limited settings than the few extra percentage points gained on GPU servers. The work demonstrates that, through judicious cross-domain architecture transfer and task-specific refinements, an extremely compressed model can remain clinically viable. We firmly believe that, for AI medical tools intended for wide deployment, breakthroughs in feasibility are as critical as pushing accuracy to its limits.

A striking finding is that UltraSeg's efficiency advantages extend far beyond the extreme-compression regime. While designed for <0.3M deployment, the same architectural principles enable 4.38M-parameter models to outperform 31M-parameter U-Net, suggesting that our design provides intrinsic representational efficiency regardless of capacity. This challenges the prevailing assumption that lightweight designs necessarily sacrifice representational power; instead, UltraSeg demonstrates that careful architectural engineering can dominate brute-force scaling across the entire parameter spectrum (0.1M–5M).

Despite the encouraging results achieved by the UltraSeg family, we are acutely aware that a performance gap remains compared with large-parameter models. We recognize that the order-of-magnitude difference in model capacity constitutes a natural and formidable upper bound, an inherent constraint that any "extreme compression" paradigm must accept. Looking ahead, we believe that performance gains in extremely lightweight models can still be pursued along complementary avenues:

First, as demonstrated in Section~\ref{parameterscaling}, moderate model scaling within a controllable range can yield further performance gains while maintaining the hard constraint of real-time inference. In practical deployment, one may increase channel widths to construct an UltraSeg family covering diverse accuracy-speed trade-offs, flexibly adapting to clinical scenarios ranging from mobile endoscopy units to standard workstations.

Second, we envision designing pre-training paradigms specifically tailored to ultra-lightweight models. Recent studies have demonstrated the benefits of self-supervised learning on colonoscopic images for downstream segmentation tasks; however, these approaches typically rely on large backbones such as ResNet or ViT. Whether such gains translate to extremely compact networks remains an open question. We hypothesize that prevailing self-supervised objectives may exhibit diminished efficacy when applied to severely capacity-constrained architectures. Consequently, we will investigate customized pre-training strategies dedicated to "extremely compressed" networks, aiming to enhance feature-extraction capabilities without inflating model size.

While standard evaluation protocols rely on static image datasets, clinical deployment requires sustained performance on continuous video streams with temporal consistency. To bridge this gap, we provide open-source implementations of real-time video segmentation pipelines (frame capture, model inference, and mask overlay) operating at $>$30 FPS on single-core CPUs. However, more comprehensive video-stream-based real-time segmentation optimization remains to be fully realized. Demonstration videos on actual colonoscopy sequences are available in the supplementary material and GitHub repository. This ensures that the reported efficiency metrics translate directly to clinical workflow compatibility.


\section{Conclusion}
\label{sec:conclusion}

This work recasts real-time polyp segmentation as an extreme-compression problem and establishes the first strong baseline under the sub-0.3M regime. Trained on mixed multi-center data, UltraSeg-108K and UltraSeg-130K achieve Dice scores of 0.7884 and 0.8038 at 256×256, with robust zero-shot generalization on ETIS-Larib and BKAI-IGH. Operating at over 50 FPS on a single CPU core, these models demonstrate the practical potential for GPU-free real-time diagnosis in primary hospitals and resource-constrained endoscopy systems.

Channel scaling experiments validate that the architectural principles are not limited to extreme compression: UltraSeg-4.38M surpasses the standard UNet baseline and approaches heavyweight state-of-the-art accuracy, confirming intrinsic representational efficiency rather than mere parameter starvation. Ablation studies further demonstrate consistent performance gains from the proposed modules across all channel configurations.

By proving that clinical-grade accuracy and real-time CPU inference are simultaneously achievable under severe resource constraints, this work removes the hardware barrier that has historically segregated high-performance medical AI from resource-limited clinical settings. It provides an immediately deployable solution for ubiquitous polyp screening and a reproducible design blueprint for real-time medical segmentation beyond colonoscopy.

\section*{Declaration of competing interest}

The authors declare that they have no known competing financial interests or personal relationships that could have appeared to influence the work reported in this paper.

\section*{Code availability}
The code for this study is publicly available at https://github.com/AI-thpremed/ultraseg.

\section*{Acknowledgements}

The authors declare that there are no acknowledgements.

\bibliography{ultraseg}
\bibliographystyle{plain}

\appendix

\end{document}